\documentclass[10pt,twocolumn,letterpaper]{article}
\pdfoutput=1

\usepackage{cvpr}
\usepackage{times}
\usepackage{epsfig}
\usepackage{graphicx}
\usepackage{caption}
\def\cvprPaperID{2202}

\cvprfinalcopy 
\begin{document}

\title{Context Augmentation for Convolutional Neural Networks}

\author{
Aysegul Dundar\\
Purdue University\\
{\tt\small adundar@purdue.edu}
\and Ignacio Garcia-Dorado\\
Purdue University\\
{\tt\small igarciad@purdue.edu}
}


\twocolumn[{%
\renewcommand\twocolumn[1][]{#1}%
\maketitle
\begin{center}
    \centering
    \includegraphics[width=0.75\textwidth]{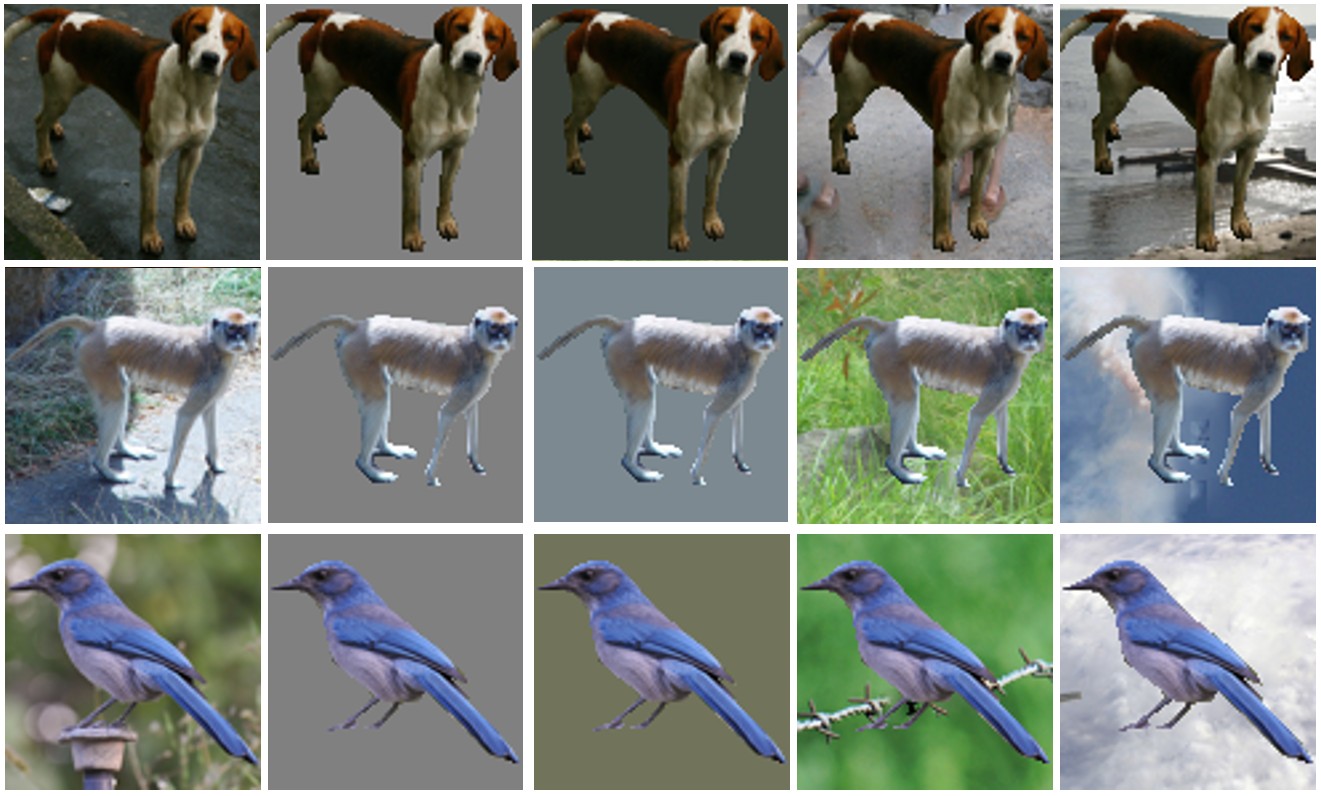}
    \captionof{figure}{Different set-up of context during training. First column has original images. Second column images are  foreground objects with uniform gray color as background. Third column has mean value of the extracted background replaced with the background.  Forth column has examples of foreground objects combined with different backgrounds from the same category as themselves. Fifth column images are foreground objects that are combined with backgrounds from other categories.}
    \label{fig:mixbg}
\end{center}%
}]

\begin{abstract}
Recent enhancements of deep convolutional neural networks (ConvNets) empowered by enormous amounts of labeled data have closed the gap with human performance for many object recognition tasks. 
These impressive results have generated interest in understanding and visualization of ConvNets.
In this work, we study the effect of background in the task of image classification.
Our results show that changing the backgrounds of the training datasets can have drastic effects on testing accuracies. 
Furthermore, we enhance existing augmentation techniques with the foreground segmented objects.
The findings of this work are important in increasing the accuracies when only a small dataset is available, in creating datasets, and creating synthetic images. 
\end{abstract}

\section{Introduction}
In recent years, ConvNets empowered by abundant amounts of labeled data have increased accuracies in many tasks, including object classification, detection, and face recognition \cite{krizhevsky_imagenet_2012, taigman2014deepface, Simonyan14c}. 
However, when the training dataset is small, the ConvNets are not able to learn the necessary features, resulting in low accuracies in these tasks.
To address this issue, several approaches such as  transfer learning, unsupervised, and semi-supervised learning have been proposed.
Nevertheless, they still require additional datasets or underperform their supervised counterparts. 

In this work, we investigate the effect of the context (i.e., background) in training of ConvNets.
500 images from a common dataset, the STL-10 dataset \cite{coates2011analysis}, have been parsed as the background and the foreground objects (i.e., the object to be classified). 
These segmented images allow us to improve the accuracy on test dataset and also enable us to understand the importance of the context. 
Additionally, we present new augmentation techniques based on the segmentation of the training dataset into background and foreground objects.

Our approach can be beneficial in several scenarios: i) Understanding ConvNets: The experiments with different backgrounds combined with foreground objects give us information about how ConvNets use background information. ii) To augment the available datasets to improve accuracies: Especially for categories, it is hard to collect data. In these cases, augmenting the data can lead to significant improvements in accuracy. iii) When a new dataset is collected. iv) For synthetic image creation; To bypass the effort of labeling and finding images, the images can be created synthetically. During this process, it is important to create backgrounds for the objects that are of interest.
In this work, we analyze how the context directly impacts the accuracy of the ConvNets. We present several results showing how the user can identify such scenarios and improve dataset creation.
	
We analyze how the background plays a very important role in the ConvNets accuracy when the dataset is of limited size. We also present new augmentation techniques and enhance other commonly utilized augmentation techniques to improve the accuracy.
Note that for our analysis, we use a very small amount of labeled data, i.e., 500 images, in order to explore how much we can advance the performance of the network when minimal data are available. The ability to train a ConvNet using a small amount of data has an advantage because data can be very expensive and for some categories large amounts of data may be difficult to find.

The main contributions of this work include:
\begin{itemize}
        \item An analysis of the context (i.e., the background) on the accuracy of a ConvNet. Using our segmented images, we analyze the importance of the background and conclude that the training dataset should contain a diverse set of backgrounds, in order to facilitate stronger accuracy in the network.  
	\item A novel augmentation technique based on the segmentation of images. When the dataset is small, the dataset can be augmented by pre-processing each image and increasing the number of input images.  We enhance the aforementioned techniques with the segmented information.
	
\end{itemize}

The results of this study are organized as follows: 
Section \ref{sec:related} discusses the related work. 
Section \ref{sec:segaug} explains the segmentation procedure.
Section \ref{sec:setupbg} and section \ref{sec:exsetup} show our experimental setup for different backgrounds and training details respectively.  
Section \ref{sec:results} presents the results.
Section \ref{sec:aug_tech} reviews augmentation techniques (2D affine transformations) and proposes segmented versions of these augmentations.
Section \ref{sec:conclusion} concludes.

\section{Related Work}
\label{sec:related}
In this section, we describe alternative techniques to address the problem of learning from few data with ConvNets such as transfer learning, and techniques related to our work such as data augmentation, synthetic dataset creation, and understanding ConvNets.

 \textbf{Learning with few labeled data:} The features that are learned from big labeled datasets also perform learned tasks well on other datasets which makes transfer learning a popular method while dealing with a small dataset \cite{razavian2014cnn, donahue2013decaf}. 
Transfer learning occurs when ConvNet's knowledge of an existing task is transferred to a new task in order to improve the networks ability to learn the new task. 
Transfer learning is applied by training a network with big labeled data and fine-tuning the classifier for the other small dataset.
Despite the success of this approach, there is still a need for big labeled datasets to train the initial network. Furthermore, the transferred features perform poorly when the datasets, tasks, are less similar \cite{yosinski2014transferable}. 
To decrease the need for big labeled datasets, extensive research has also been dedicated to unsupervised and semi-supervised learning algorithms which also utilize unlabeled data (e.g. \cite{dosovitskiy2014discriminative}.
However, these approaches underperform their supervised counterparts when the amount of labeled data is large, which shows that they do not efficiently represent the necessary features. 
	
\textbf{Synthetic Dataset:} One approach to bypass the need for creating a labeled dataset manually is to create the images synthetically.
	Jaderberg et al. \cite{jaderberg2014synthetic} successfully use a synthetic dataset for Optical Character Recognition (OCR) systems.
    In order to generate the synthetic images, they randomly select fonts and render them with different colors over a background.
    This process, known as background blending, is especially important for scene text recognition in synthetic OCR systems. This process is an improvement to traditional OCR techniques which fail at detecting text in scene images due to the fact that they are tuned to work on black-and-white text and line-based printed documents.
Thus, to create a synthetic dataset for generic object recognition, a more elaborate scheme is necessary for the background than was utilized in traditional OCR systems.

Recently, Peng et al. \cite{peng2014exploring} use 3D CAD models to create a synthetic dataset for object detection.
While they perform experiments with different backgrounds on their detector, their experiments are built on a pretained ConvNet \cite{krizhevsky_imagenet_2012} on the Imagenet dataset \cite{deng2009imagenet}.
Therefore, they do not show the importance of having a variety of backgrounds while training an object classifier. 

\textbf{Data Augmentations:}  Data augmentation is the technique which improves the accuracy of ConvNets by increasing the size of the training dataset.
Most augmentation techniques perform a 2D local transformation on the image (e.g., rotation and translation) to generate multiple variations of one input image~\cite{ciresan2012multi}.
This technique is commonly used to learn invariant features \cite{wu2015deep, ciresan2012multi, krizhevsky_imagenet_2012}.
To facilitate augmentation, Dosovitskiy et al.~\cite{dosovitskiy2014discriminative} use a single image (i.e., a seed) per class to create an augmented dataset. Each image belongs to a different category and is accompanied by hundreds of transformed versions of itself (e.g. color, contrast).
The network learns discriminative features that are invariant to some typical transformations. 
After the network is trained, the classifier is retrained by the labeled samples of the recognition task. 
\begin{figure}[t]
   \centering
   \includegraphics[width=\columnwidth]{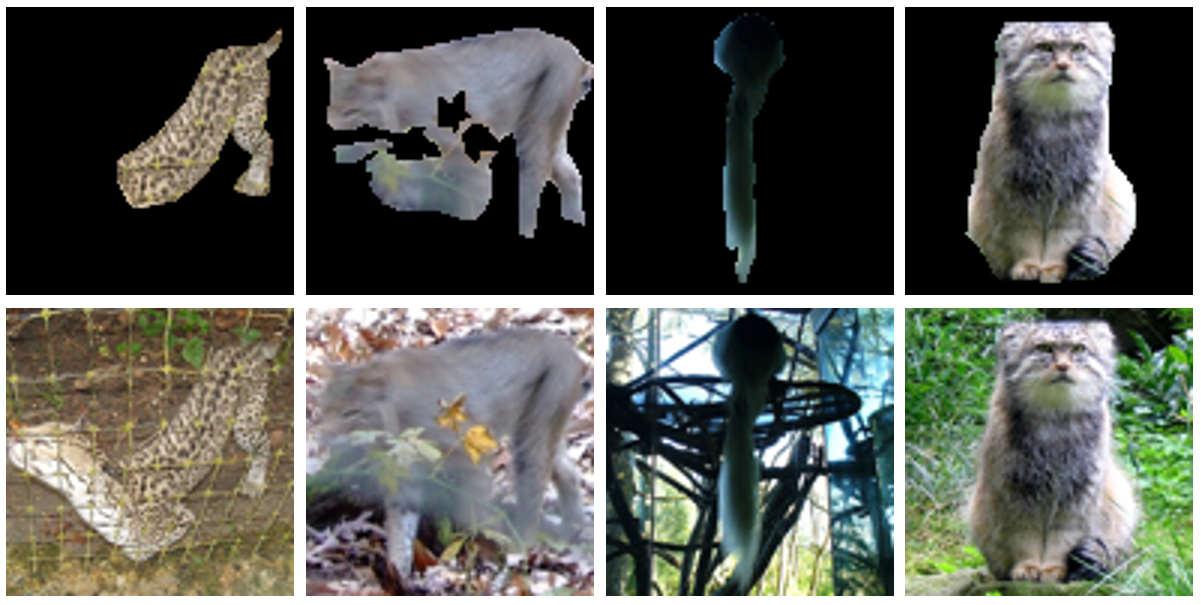}
   \caption{
   The foreground segmented images in the first row and the original image in second.
   All images belong to the cat category.}
   \label{fig:context}
\end{figure}

\textbf{Understanding ConvNets:} With the increasing success of ConvNets, there has been a growing interest in understanding various dimensions of its behavior such as the training process and the features that are learned.
Several visualization techniques have been proposed to analyze and investigate what kind of features are learned in different layers of ConvNets \cite{mahendran2014understanding, zeiler2014visualizing, springenberg2014striving}.  
There has also been further effort to measure the encoded invariance of a ConvNet: invariance to 2D transformations \cite{lenc2014understanding} and low-level cues \cite{peng2014exploring}. 

Additionally, there have been studies to analyze which inputs can fool the ConvNets. For example, the study conducted by Szegedy et al.  \cite{szegedy2013intriguing} shows that small perturbations in the input images can make the network alter its prediction about the image from a correct label to a wrong one.
Furthermore, Nguyen et al. \cite{nguyen2014deep} show examples of images that are unrecognizable to humans, but are categorized by state-of-the-art deep ConvNets as familiar objects with certainties greater than $99.6\%$.
These findings raise questions about the generality, or the ability of ConvNets to be tolerant to small changes in images.

\section{Segmentation-based Augmentation}
\label{sec:segaug}
In this section, we describe the importance of image context, how images are segmented, and how we use the segmented images to create a novel augmentation technique. 

\subsection{Image Context}
The context of an image (i.e., the background) has significant importance for us while recognizing objects, especially when images are small. 
Figure~\ref{fig:context} shows the foreground in the first row (i.e., the main object to classified) and the original image in second. 
According to our observations, it is hard to recognize some objects using just the foreground information because the context of the image helps to classify them.
The key inspiration of adding the dimension of foreground and background in object recognition comes from how we perceive the world.
Most animals have binocular vision, which enables them to perceive depth, and to distinguish objects by differentiating the object in respect its surrounding context.
In the following section, we analyze how the extra information added by segmentation can be used to augment the dataset and improve the accuracy.

\subsection{Segmentation}
\begin{figure}[t]
   \centering
   \includegraphics[width=\columnwidth]{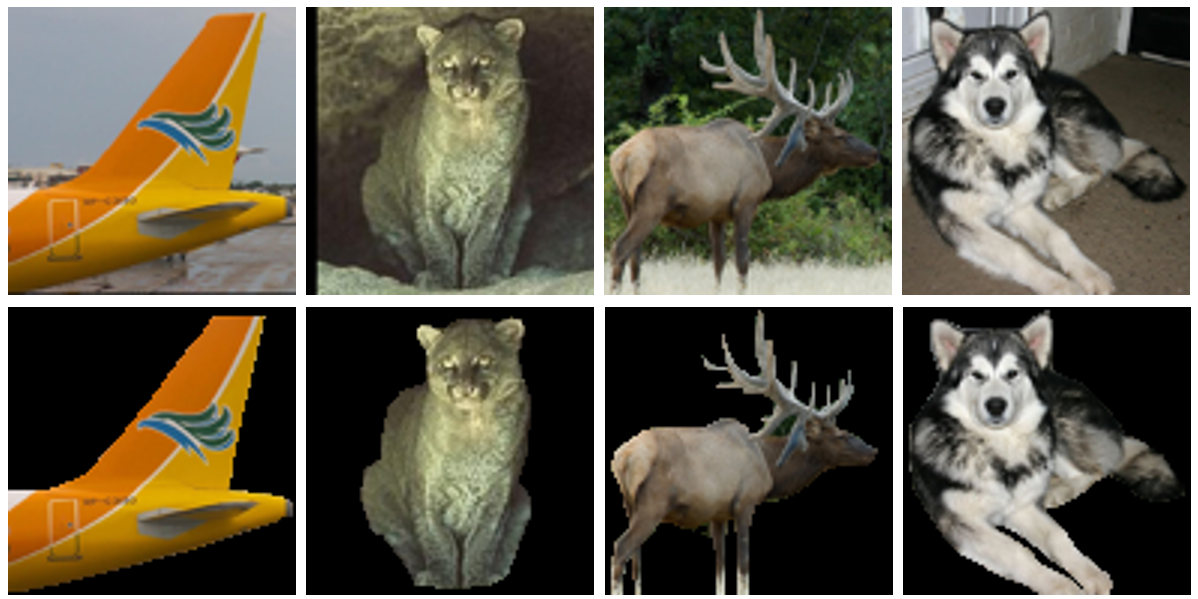}
   \caption{Examples of segmented images.}
   \label{fig:segmentation}
\end{figure}

To be able to analyze the importance of the context, we need to discriminate between foreground and background.
We explore several options to segment the foreground objects: i) datasets with depth data, ii) automatic segmentation, iii) manual segmentation.
We discarded datasets that contain depth information because they are very limited in size, and it is hard to compare with other well-known datasets. 
We expect that these datasets will become more abundant with the development of depth cameras in mobile environments (\cite{ng2005light,martinellodual}).
We explored the automatic segmentation approach using the methods proposed by Long et al.~\cite{long2015fully} and Comaniciu et al.~\cite{comaniciu2002mean}, as well as by using a graph-cut-based segmentation~\cite{rother2004grabcut}. 
However despite their strong performance on other datasets, these algorithms perform poorly on the STL-10 dataset.
The reason for that can be that the images are small $96\times96$ and are also noisy. 
We decided to pursue a manual segmentation. This has two main advantages: One, the segmentation does not contain noise that would propagate to the augmentation, and two, it allows us to use a well-known dataset.

To speed up the process of creation and save additional information like the click positions over time as well as the final mask, we developed an interactive segmentation application. We realized that we needed to present the same image in several sizes to the users for the users to segment the objects properly. Several examples of the segmentations can be seen in Figure~\ref{fig:segmentation}.

\subsection{Segmentation Augmentation}

\begin{figure}
   \centering
   \includegraphics[width=\columnwidth]{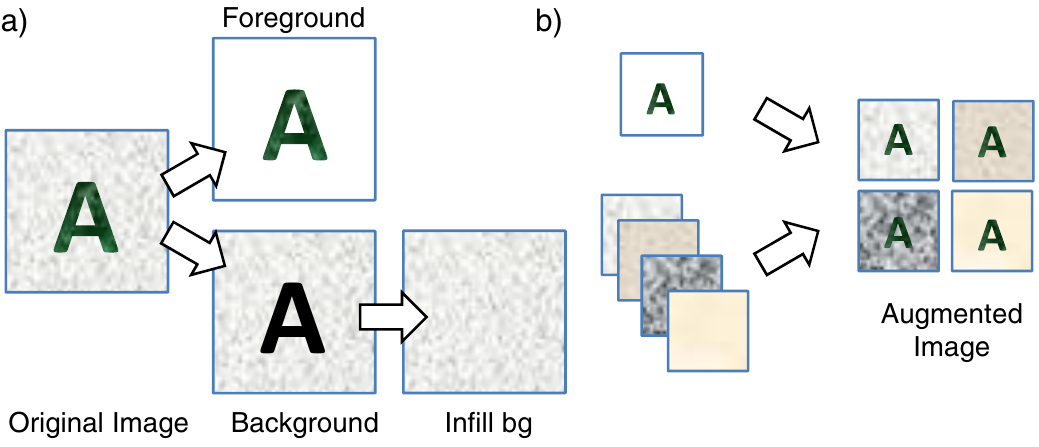}
   \caption{Segmented-based Augmentation Overview. a) The image is split into foreground and background. The background is filled in. b) Each foreground can now be combined with the infilled backgrounds which creates new examples for the training.}
   \label{fig:schematic}
\end{figure}

Once the images are segmented, we can alter their background to analyze their importance and create our segmentation-based augmentation (Figure~\ref{fig:schematic}). Note that since the foreground of each image does not match the others, we need an algorithm to fill in the gaps that have occurred because of the removal of segmented objects from the images. Recently there have been several approaches to completing the scene of a photograph. Hays et al.~\cite{Hays2007} patch up holes in images by exploiting a large database of pictures with similar statistics. Other studies focus on the minimization of artifacts of pattern-based inpainting~\cite{Daisy2013}. 

\begin{figure}
   \begin{center}
   \includegraphics[width=\columnwidth]{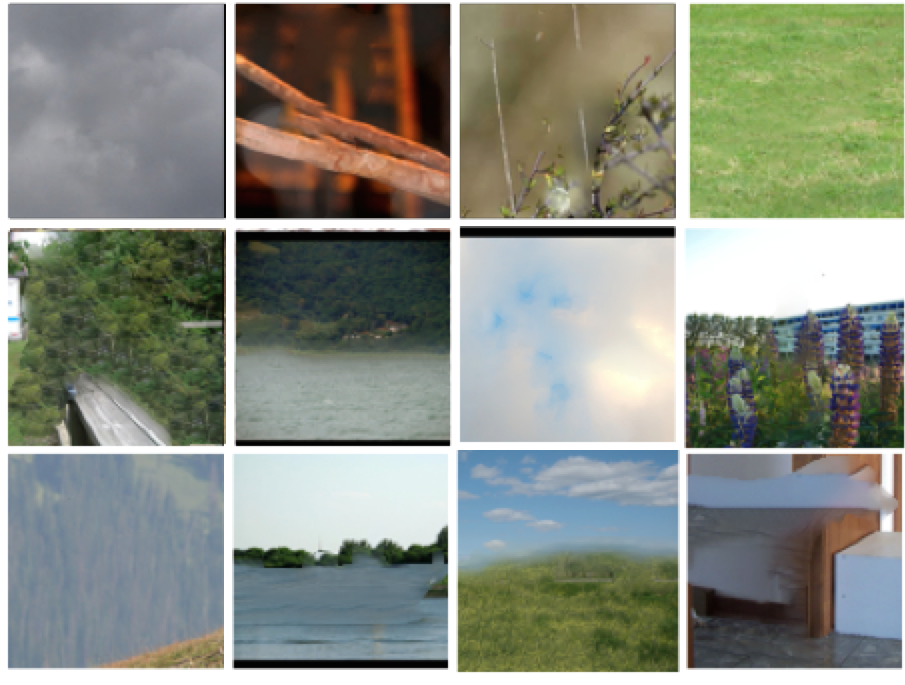}
   \end{center}
   \caption{Images of extracted backgrounds. Gaps are automatically filled using  \textit{patch-match} approach \cite{barnes2009patchmatch}.}
   \label{fig:foreground}
\end{figure}

For our task, we use the method proposed by Barnes et al.~\cite{barnes2009patchmatch}. This algorithm quickly finds correspondences between small square regions of an image. Using corresponding patches extracted from its surroundings, the algorithm infills gaps in images. Some examples of infilled background images can be seen in Figure \ref{fig:foreground}.

\section{Set-up with Backgrounds}
\label{sec:setupbg}
In this section, we present different combinations of backgrounds and foreground objects we use in our experiments. They are as follows:

\begin{enumerate}
\item \textbf{ Only bg} (only background): After the foreground objects are segmented and removed from the background, the gaps are filed with \textit{patch-match} approach\cite{barnes2009patchmatch}.
Examples of the backgrounds that are created with this process are shown in Figure \ref{fig:foreground}.
Training with these images is interesting because even though they do not include the objects in which we are interested, they have the information of the environment in which the object is normally found.             	                
\item \textbf{Gray bg with fg} (gray background with the segmented foreground objects): These images have the segmented foreground objects while their backgrounds are replaced with a gray value.
This process takes all the information about the background and leaves only the foreground objects.  Examples can be seen in Figure \ref{fig:mixbg} (second column).             	
\item  \textbf{Mean value of bg with fg} (mean value of the removed background with the segmented foreground): These images instead of a uniform gray color as background have the mean value colors of the extracted backgrounds.
Therefore, if the image was a plane in the air, the mean value is most likely to be blue, and then the background becomes a uniform blue.
Examples of these images can be seen in Figure \ref{fig:mixbg} (third column). 
\item  \textbf{Bg same category with fg} (backgrounds from the same category combined with the foreground objects from the same category): In this set-up, segmented foreground objects from each category are combined with the background images from the same category as well. 
Examples can be seen in Figure \ref{fig:mixbg} (fourth column), in the first row a segmented dog from the first image is combined with a background image from image that belongs to a dog category.
These images preserve the `correct label' for the backgrounds and the foreground objects.
For segmented 500 images, 50 images per category, this process creates $50\times50\times10=25000$.
\item  \textbf{Bg all categories with fg} (backgrounds from all the categories combined with each foreground object): These images are basically all possible combinations of the segmented foreground objects and the background images.
For segmented 500 images, this process creates $500\times500=250000$.
Two of the examples can be seen in the last column of Figure \ref{fig:mixbg}, the first one is an image of a dog with a background from an image of a plane. The second one is a monkey with the background again that belongs to a plane. These images look less realistic than the previous examples to human observers because we know that a monkey can not hang in the air, and there will not be a similar image in the test dataset.
Combining the images as such provides many examples for each category. Therefore, it may decrease the problem of the network overfitting to the training dataset.
On the other hand, it removes the background information in a way that the network cannot take advantage of the background while categorizing an object because the same background appears for each category.
\end{enumerate}
\section{Experimental Setup}
\label{sec:exsetup}
We use the STL-10 dataset for our experiments \cite{coates2011analysis}.
This dataset contains 5000 training and 8000 testing images from 10 categories.
In our experiments, we use 500 of these training images, 50 from each category.
The images are RGB colors and $96\times96$.
We only pre-process the images by global contrast normalization.

In the experiments, we use a 4 layer ConvNet configured as (96) 7c- 3p - (256) 5c - 2p - (512) 3c - 2p - (10) c1 where (96) 7c denotes convolution with 96 filters each is $7\times7$.
3p  denotes $3\times3$ pooling with a stride of 3.
ReLU non-linearity operation follows each convolution layer.
As a classifier, global average pooling is used \cite{lin_2013_nin} followed by a softmax layer.

The networks are trained with a learning rate of 0.1 and a momentum of 0.9. 
The training used stochastic gradient descent algorithm with a batch size of 10.
For each experiment, we repeat the training 10 times with different random seeds and provide mean value and standard deviations of the test accuracies.
For the experiments that each foreground objects may appear with many different backgrounds, we make sure that in the same epoch (500 unique images) each foreground object and background appear only once with random combinations.

\section{Results on the Background}
\label{sec:results}
The 500 training images without any augmentation results in the accuracy of  $47.42\pm 1.14 \%$. 
With the  500 segmented foreground images and their filled background counterparts, we perform many experiments as shown in Table \ref{table:bg_test}.

\textbf{Only bg:}  In our first experiment, we use only the background for training. Examples are shown in Figure \ref{fig:foreground}.
We train the network with these images by conserving the labels of the original images.
Even though we have removed the foreground objects (i.e., the objects of the categories) from the examples, we are experimenting to see if the network can still learn something from the background that would result in a test accuracy that is better than chance ($10\%$).
Ships are usually in the water, planes are in the sky, and birds are on the branches of trees. Therefore, it would not be surprising if the network achieves test accuracy better than chance.
In fact, the network correctly classifies $31.3\%$ of the images from the test dataset which is quite surprising.

\begin{table}
\normalsize
  \caption{Experiments with different combination of backgrounds (bg) with the foreground objects (fg).}
\centering
   \begin{tabular}  {| c | c | c | }
      \hline
      
                      		                         &\#images			&Accuracy			\\ \hline
      \hline
       Original images                 	         &$500$		         & $47.4\pm 1.1 \%$		 \\ \hline
       Only bg                 	                 &$500$		         & $31.3\pm 0.8 \%$		 \\ \hline
       Gray bg with fg                     	 &$500$		         & $28.7\pm 1.0 \%$		 \\ \hline
       Mean value of bg with fg              &$500$		         & $36.7\pm 0.9 \%$		 \\ \hline
       Bg same category with fg            &$25000$		 &$54.4\pm 0.9 \%$		\\ \hline
       Bg all categories with fg               &$250000$		 &$48.5\pm 1.0 \%$		 \\ \hline
   \end{tabular}
   \\
   \label{table:bg_test}
\end{table}

\textbf{Gray bg with fg:}  In the second experiment, we use the foreground objects with a gray background.
This removes a lot of information from the training dataset. Therefore, the network performs poorly on the test dataset.
In fact, it gives worse performance than the training with only background images ($28.7\%$ as opposed to $31.3 \%$).

\textbf{Mean value of bg with fg:}  In the third experiment,  the gray value is replaced with the mean color of the background as can be seen in Figure \ref{fig:mixbg}-third row. It provides more information and variety in the training dataset.
The mean value is also partially adding a background information.

\textbf{Bg same category with fg:}  In our fourth experiment, we increase the dataset by combining the foreground objects with the backgrounds from the same categories.
Examples can be seen in Figure \ref{fig:mixbg} (fifth column). This combination preserves the correct label for both the foreground images and the background images while creating 25000 examples.
Another surprising result of this experiment is that the accuracies increase $15\%$ compared to the training with original images.

\textbf{Bg all categories with fg:} In the fifth experiment, we combine each foreground object with each background object. Segmentation of 500 images gives us 500 unique foreground and background images. Combining them creates 250000 different images.
The result is slightly better than the training with original images ($48.5\%$ as opposed to $47.4\%$).
This experiment is also in a way removing the background information from the dataset.
Because the same background appears for each category, the network cannot take advantage of the statistical information of the background while classifying an object during training.
From this perspective, it is similar to the experiment with gray backgrounds.
On the other hand, it also creates many examples of the foreground objects in different backgrounds and reduces the overfitting problem.

These experiments show us the importance of the context while training ConvNets.
As they learn the foreground images, they also learn the backgrounds.
Interestingly, as evidenced in our first experiment, sometimes background information is all that is needed to categorize some images.
Another interesting finding is the significant increase in accuracy when the data are augmented with the combinations of foreground and background images from each category.

Finally, our last experiment shows the delicate balance between  having more examples which can potentially increase accuracy and removing background information in order to create these examples which decreases accuracy. 
The result is better than the training with original images, but is much worse than the training with the combinations of the same label of foreground and background images.
On the other hand, even though the results are worse, we know that the network uses the foreground objects to recognize the categories.
This recognition can be useful for many applications and is still better than just using the original images, because despite the fact that the original images use background information, they still result in lower accuracy.

\section{Augmentation Techniques}
\label{sec:aug_tech}

\begin{figure*}[!h]
   \begin{center}
   \includegraphics[width=\textwidth]{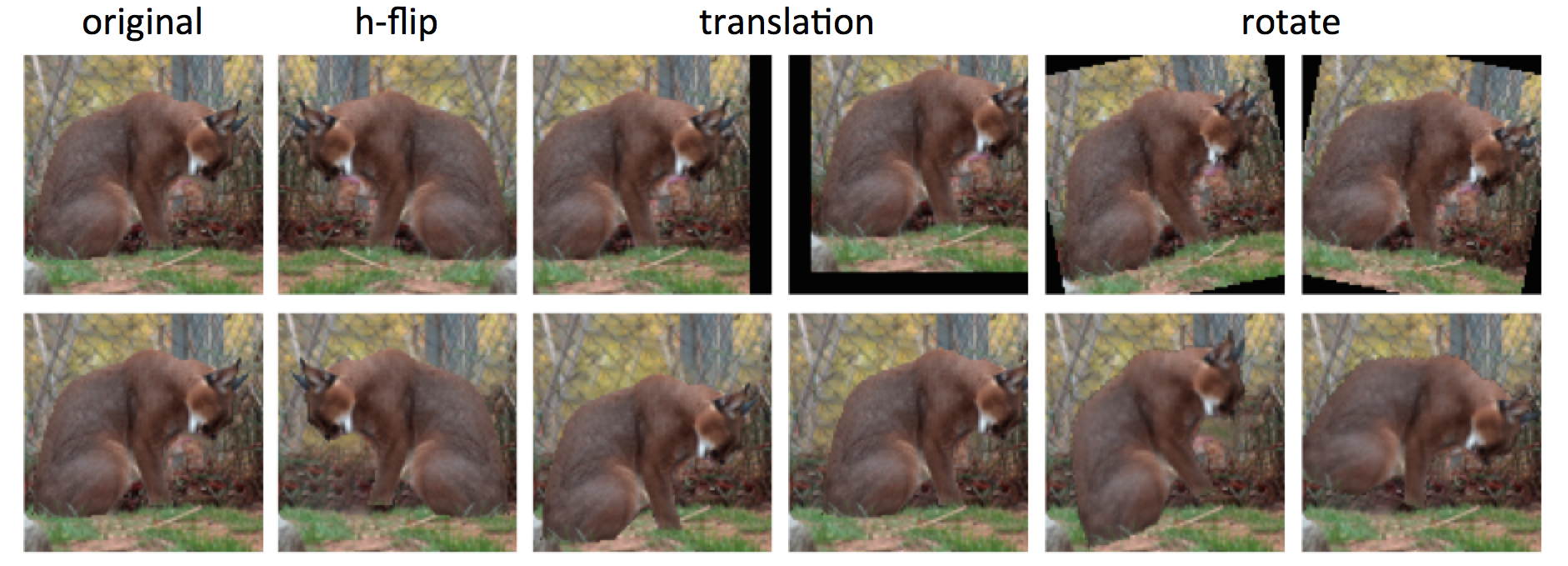}
   \end{center}
   \caption{First row are images that standard data augmentations applied to the whole image. Second row are the images where the backgrounds stay the same and the augmentations are only applied to the foreground objects.
   They are referred  to as segmented versions of the corresponding data augmentations.}
   \label{fig:augs_ex}
\end{figure*}

We are interested in understanding if the segmented objects and background can be useful for data augmentation that would result in increased accuracy. 
In the previous section, we saw that combining foreground and background images can provide an increase in accuracy.
In this section, we experiment with 2D transformations of the segmented objects.

First, we review standard data augmentation approaches that are used to train ConvNets. 
We are studying how each augmentation technique improves the test accuracies when the amount of labeled data is very few.
In addition, we propose new augmentation techniques that are based on  segmented object images and background images.
Examples of the augmentation techniques we used in our experiments can be seen in  Figure \ref{fig:augs_ex} where the first row contains examples of standard augmentation methods and the second row shows examples of segmented version of these augmentations.
They are as follows:

\begin{itemize}
\item Horizontal flipping (hflip):
Images are flipped horizontally, which gives us the mirror images of the originals.
Segmented horizontal flip is when only the the foreground objects are flipped while the background stays the same (Figure \ref{fig:augs_ex} - second column).

\item Translation: Images are shifted  of random values which are uniformly selected between $(-20, 20)$ pixels in vertical and horizontal directions.
Segmented translation of the images corresponds to the same operation but only applied to the segmented foreground object with the values of $\{-20, -15, -10, -5, 5, 10, 15, 20\}$ (Figure \ref{fig:augs_ex} - third and forth column)).

\item Rotation: Images are rotated by a random angle between $(-10, 10)$.
Segmented rotation of the images, similar to the previous examples, rotation is only applied to the foreground objects with angles of $\{{-10}, {-5}, 5, 10\}$ (Figure \ref{fig:augs_ex} - fifth and sixth column)). 
\end{itemize}

We also experimented with several other data augmentation methods in addition to the above (e.g. changing the color - adding a value to the hue component in HSV representation, and changing the contrast of the whole images,
changing the color of the foreground objects and the background independently and adding Gaussian Noise), but we did not observe any accuracy gain with these augmentations.

In our first experiment in this section, we test standard data augmentations on the original dataset.
We run experiments in 2 scenarios: 500 training images and 5000 training images (all labeled examples from the STL-10 dataset).
The network achieves an accuracy of $47.4\%$ with 500 training images, and it achieves an accuracy of $69.8\%$ with 5000 training images, without any augmentation techniques applied.

\begin{figure}
   \centering
   \includegraphics[width=\columnwidth]{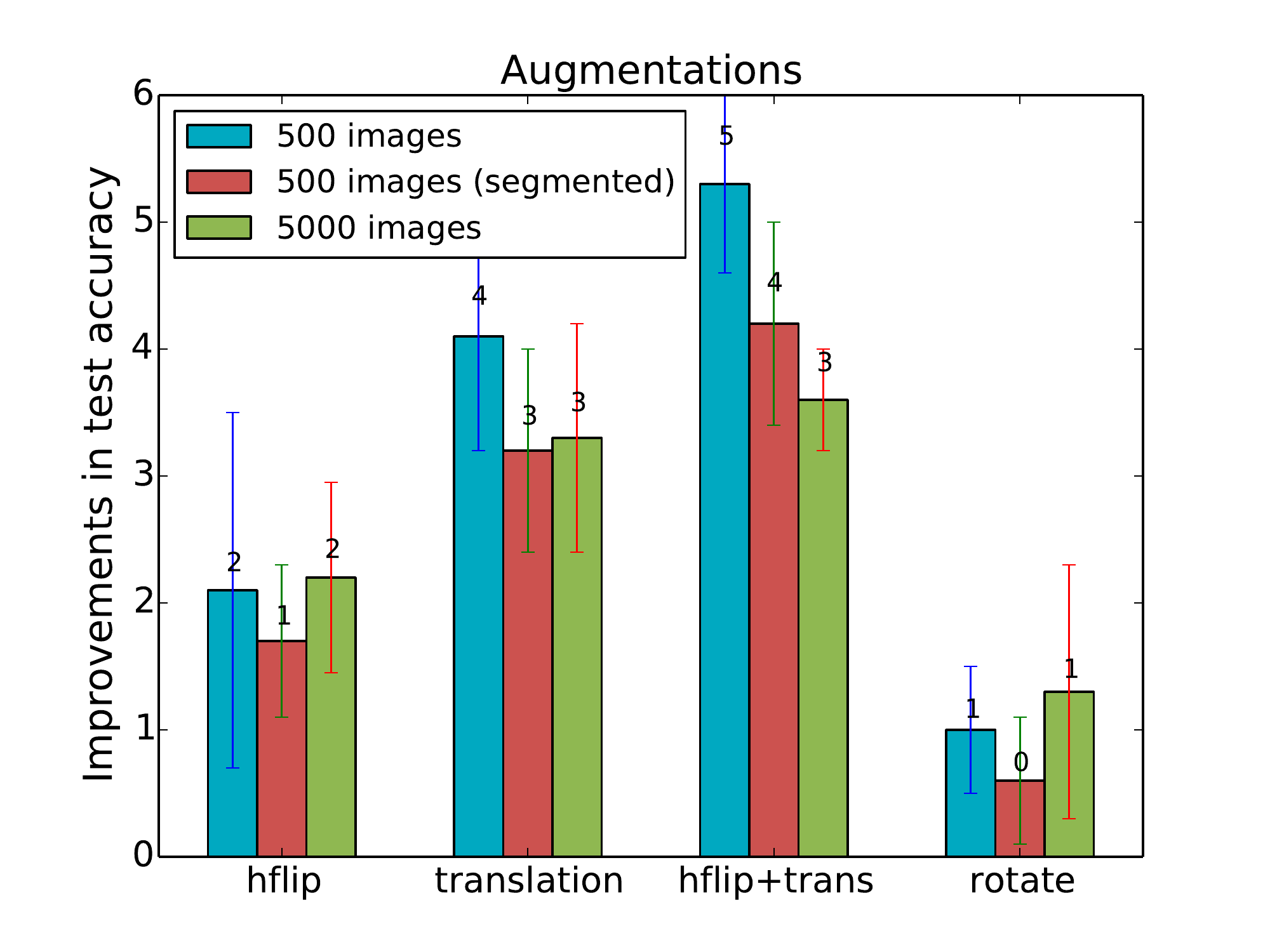}
   \caption{Influence of augmentations on the STL-10 dataset when 500 images and 5000 images are used. Baseline accuracy value for the training with 500 images is $47.42\%$ and with 5000 images, it is $69.8\%$.
   For 500 images, the segmented versions of the augmentations results (label 500 images segmented) are given as well as the standard augmentation techniques' (label 500 images).
   For 5000 images, only the standard augmentations are applied (label 500 images).}
   \label{fig:augs}
\end{figure}

Figure \ref{fig:augs} displays the influence of each data augmentation on the test accuracy.
The segmented augmentation versions increase accuracy less than their original counterparts.
In the segmented augmentations, the background stays the same. In standard augmentation the background is also augmented which further increases variety in the training dataset.
On the other hand, the segmented and the original augmentation techniques create different images, and they can be used together to further boost the performance.

In our final experiments, we combine all the techniques that result in improvements when training the ConvNet with 500 images.
Table \ref{table:bg_test} presents the results.
Each row includes data augmentations from the previous rows.
Note that when combined with other augmentation techniques, rotation did not improve results. Therefore, rotation has been removed from the table.
With only 500 images enhanced with the augmentation techniques, the network achieves an accuracy of $59.5\pm 0.8 \%$.
The cumulative augmentation and segmentation algorithms applied, resulted in a $25.5\%$ improvement in image recognition from the initial dataset.

\begin{table}[!h]
\normalsize
   \caption{Accuracies with 500 unique examples. Each row includes the augmentation from the previous rows.}
\centering
   \begin{tabular}  {| c | c | }
      \hline
      
                      		                         &Accuracy			\\ \hline
      \hline
       Original images                 	         & $47.4\pm 1.1 \%$		 \\ \hline
       Bg same category with fg            &$54.4\pm 0.9 \%$		\\ \hline
       Horizontal flip                              &$55.4\pm 0.7 \%$           \\ \hline
       Translate                                     &$58.9\pm 0.4 \%$           \\ \hline
       Segmented hflip                          &$59.5\pm 0.8 \%$                \\ \hline
   \end{tabular}
   \\
   \label{table:bg_test}
\end{table}

\section{Summary}
\label{sec:conclusion}

We provided an analysis of the impact of background while training ConvNets.
We also proposed different techniques to improve the accuracies when only a small dataset is available to us.

Recently, deep ConvNets have closed the gap with human performance in many tasks and contain promising advances in the field of visual understanding.
The problem is their need for huge labeled datasets.
The question which arises is whether the ConvNets fully utilize the data provided to them.
In this work, we observed that creating datasets which utilize foreground and background objects from the same category increased the performance of the network.

We also observed that in certain instances  background may give enough information for the network to categorize some of the objects while in other instances the network is unable to do so with the given background information. 
Furthermore, we proposed new augmentation techniques that provide accuracy increase in ConvNets.
The findings of this work are important in increasing the accuracies when only a small dataset is available, in creating datasets, and creating synthetic images.
The ability to train a ConvNet using a small amount of data has an advantage because data can be very expensive and for some categories large amounts of data may be difficult to find.


{\small
\bibliographystyle{ieee}

}


\end{document}